\documentclass[aip, preprint]{revtex4-1}

\newcommand{\vv}[1]{\textbf{#1}}
\usepackage{amsmath}
\usepackage{amsfonts}
\usepackage{amssymb}
\usepackage{graphicx}
\usepackage{url}
\draft 

\begin{document}


\title{Using Data Assimilation to Train a Hybrid Forecast System that Combines Machine-Learning and Knowledge-Based Components} 


\author{Alexander Wikner}
\thanks{A. Wikner and J. Pathak contributed equally to this work. Correspondence should be addressed to: awikner1@umd.edu}
\affiliation{Department of Physics, University of Maryland, College Park, MD 20742}

\author{Jaideep Pathak}
\thanks{A. Wikner and J. Pathak contributed equally to this work. Current address: Lawrence Berkeley National Laboratory (LBNL), Berkeley, CA 94720}
\affiliation{Department of Physics, University of Maryland, College Park, MD 20742}

\author{Brian R. Hunt}
\affiliation{Department of Mathematics, University of Maryland, College Park, MD 20742}
\affiliation{Institute for Physical Science and Technology (IPST), College Park, MD 20742}

\author{Istvan Szunyogh}
\affiliation{Department of Atmospheric Sciences, Texas A\& M University, College Station, TX 77843}

\author{Michelle Girvan}
\affiliation{Department of Physics, University of Maryland, College Park, MD 20742}
\affiliation{Institute for Physical Science and Technology (IPST), College Park, MD 20742}
\affiliation{Institute for Research in Electronics and Applied Physics (IREAP), College Park, MD 20742}

\author{Edward Ott}
\affiliation{Department of Physics, University of Maryland, College Park, MD 20742}
\affiliation{Institute for Research in Electronics and Applied Physics (IREAP), College Park, MD 20742}
\affiliation{Department of Electrical and Computer Engineering, University of Maryland, College Park, MD 20742}


\date{\today}

\begin{abstract}
We consider the problem of data-assisted forecasting of chaotic
dynamical systems when the available data is in the form of noisy
partial measurements of the past and present state of the dynamical
system. Recently there have been several promising data-driven
approaches to forecasting of chaotic dynamical systems using machine
learning. Particularly promising among these are hybrid approaches
that combine machine learning with a knowledge-based model, where a
machine-learning technique is used to correct the imperfections in the
knowledge-based model.  Such imperfections may be due to incomplete
understanding and/or limited resolution of the physical processes in
the underlying dynamical system, e.g., the atmosphere or the ocean.
Previously proposed data-driven forecasting approaches tend to
require, for training, measurements of all the variables that are
intended to be forecast. We describe a way to relax this assumption by
combining data assimilation with machine learning.  We demonstrate
this technique using the Ensemble Transform Kalman Filter (ETKF) to
assimilate synthetic data for the 3-variable Lorenz system and for the
Kuramoto-Sivashinsky system, simulating model error in each case by a
misspecified parameter value. We show that by using partial
measurements of the state of the dynamical system, we can train a
machine learning model to improve predictions made by an imperfect
knowledge-based model.
\end{abstract}


\maketitle 
\begin{quotation}
A common desired function is that of cyclic forecasting of a chaotic system's state, in which a set of forecasts to various time increments in the future is periodically produced once every cycle time $T$ (e.g., for weather forecasting, $T=6$ hours is common). To do this, at the beginning of each cycle, available measurements taken during the last cycle are combined with (``assimilated into'') a forecast of the current state made at the beginning of the previous cycle to produce an estimate of the current system state. This estimate then serves as the initial condition for the prediction model which produces the next set of forecasts, and so on. The process of cyclic formation of the required initial conditions is called ``Data Assimilation.'' Recently, prediction models utilizing machine learning have been developed and shown to have great promise. Thus, the training and use of machine-learning-assisted prediction models in cyclic forecasting systems is of considerable interest. To enable this, it is
necessary to train the machine-learning system, using measured data,
to forecast variables that are not measured.  Here we show this can be accomplished using data assimilation, and demonstrate the
ability of machine learning to improve the forecast of
variables for which no directly measured training data is available.
\end{quotation}
\section{Introduction}
The application of machine learning (ML) to forecasting chaotic dynamical
systems (e.g., [\onlinecite{lorenz1963deterministic}]) has been a subject of much recent interest [\onlinecite{jaeger2004harnessing, lukovsevivcius2012practical, pathak2017using, pathakmodel, pathak2018hybrid, bocquet_data_2019, chattopadhyay_data-driven_2019, brajard2020combining, bocquet_bayesian_2020, wikner_combining_2020, arcomano_machine_2020, bocquet2020online}]. Promising results of tests promote the possibility of using ML for large scale applications such as weather forecasting. Particularly, Ref.~\onlinecite{pathakmodel} indicates that machine-learning techniques can be scaled effectively to forecast arbitrarily high-dimensional chaotic attractors. In cases like weather forecasting where a skillful knowledge-based model is available, hybrid approaches that use ML to correct model deficiencies are especially promising [\onlinecite{pathak2018hybrid, reichstein_deep_2019}]. By ``knowledge-based'', we mean derived from fundamental principles such as the laws of physics.

So far, most data-driven forecasting approaches assume that training data is available for all the variables to be forecast. This assumption is relaxed in [\onlinecite{bocquet_data_2019,brajard2020combining,bocquet_bayesian_2020,bocquet2020online}], which describe techniques to develop a data-driven forecast model on a specified, but potentially only partially measured, model state space. Using an approach different from those in Refs.~\onlinecite{bocquet_data_2019,brajard2020combining,bocquet_bayesian_2020,bocquet2020online}, in this paper, we develop a technique for hybrid knowledge-based/ML systems in the partially measured scenario.
Both our hybrid approach and the purely data-driven approaches in [\onlinecite{bocquet_data_2019,brajard2020combining,bocquet_bayesian_2020, gmd-2020-211}] are based on the paradigm of data assimilation (DA), which has long been used in weather forecasting and other applications.

When a forecast model is available, DA (e.g., [\onlinecite{law_data_2015,asch_data_2016,fletcher_data_2017}]) determines initial conditions for an ongoing sequence of forecasts, based on an ongoing sequence of measurements. DA does not require that current measurements are sufficient to determine the current model state; instead, it typically uses the forecast model to propagate information from past measurements to augment current information. This technique helps both to constrain the model state and to filter noise in the measurements.

Ideas from ML and DA have been combined in various ways in the literature. DA has been used with data-driven forecast models based on delay-coordinate embedding [\onlinecite{hamilton_ensemble_2016}] and analog forecasting [\onlinecite{lguensat_analog_2017}]. On the other hand, with appropriate training data, machine-learning techniques can be taught to perform DA, without explicitly formulating a forecast model [\onlinecite{lu_reservoir_2017}]. DA has also been employed as a paradigm and a tool for the training of machine-learning systems [\onlinecite{puskorius_neurocontrol_1994,abarbanel_machine_2018,bocquet_data_2019,brajard2020combining,bocquet_bayesian_2020,bocquet2020online}].

We give a brief introduction to DA in Sec.~\ref{sec:dataAssimilation}. 
In Sec.~\ref{sec:mletkf-algorithm}, we formulate a potentially effective scheme combining DA (Sec.~\ref{sec:dataAssimilation}) with machine-learning-assisted prediction (as described in Ref. ~\onlinecite{pathak2018hybrid}) to create a cyclic forecast system. In Sec.~\ref{sec:iterated-mletkf}, we propose an iterative procedure to improve the DA cycle and potentially produce improved ML-assisted forecast models. In Sec.~\ref{sec:results}, we show the results of numerical experiments.

\section{Methods}\label{sec:dynamical}
We consider the problem of forecasting a chaotic, nonlinear dynamical system represented by the equation
\begin{align}\label{eq:dynamical}
\frac{{d\tilde{\vv{x}}}(t)}{dt} &= \vv{F}[\tilde{\vv{x}}(t)].
\end{align}
We assume that we are only able to obtain partial measurements of the state $\tilde{\vv{x}}$ at regular time intervals. We denote the measurements made at time $k\Delta t$ by $\vv{y}_k$, so that
\begin{align}\label{eq:measurement}
\vv{y}_k &= \vv{H}\tilde{\vv{x}}(k\Delta t) + \eta_k.
\end{align}
In Eq.~(\ref{eq:measurement}), $\vv{H}$ is often called the observation operator; here, we call it the measurement operator. For our exposition, we assume $\vv{H}$ to be linear, as will be true in the numerical experiments performed in this paper. (Many DA methods allow a nonlinear measurement operator, and our ML method is compatible with this case.) The random variable $\eta_k$ represents measurement noise, which is typically assumed to be Gaussian with mean $\vv{0}$ and covariance matrix $\vv{R}$.

In applications, we do not have full knowledge of the dynamics of the system, and, as such, Eq.~(\ref{eq:dynamical}) is unknown. What often is available is an imperfect model of the dynamics, with model error due to imperfect representations of some of the physical processes that govern the system. We assume that we have access to such an imperfect ``knowledge-based'' model denoted by $\vv{G}$ and that the dynamical equation,
\begin{align}\label{eq:imperfect}
\frac{{d\vv{x}}(t)}{dt} &= \vv{G}[\vv{x}(t)],
\end{align}
can, via evolution of $\vv{x}$, be used to forecast an approximation of the future $\tilde{\vv{x}}$-dependent state properties of interest. For our exposition below, we assume that $\vv{x}$ and $\tilde{\vv{x}}$ lie in the same space, but in an application such as weather forecasting, $\vv{x}$ can represent the projection of the physical state $\tilde{\vv{x}}$ onto a model grid.

\subsection{Data Assimilation}\label{sec:dataAssimilation}
In DA, the initial state for Eq.~(\ref{eq:imperfect}), $\vv{x}(0)$, is estimated from the current measurement $\vv{y}_0$ and all past measurements $\lbrace \vv{y}_{k} \rbrace_{-T \leq k<0}$, using the model, Eq.~(\ref{eq:imperfect}), to imperfectly propagate information from the past to the present. Many DA methods are based on the following nonlinear least-squares problem. Given a dynamical model, Eq.~(\ref{eq:imperfect}), and a set of measurements $\lbrace \vv{y}_k\rbrace_{-T \leq k \leq 0}$, find a trajectory $\vv{x}(t)$ of the model $\vv{G}$ that minimizes the cost function given by
\begin{align}\label{eq:dataAssimilationbase}
J_0(\vv{x}(t)) = \sum_{k = -T}^0 \rho^{k} \left[\vv{y}_k - \vv{H}\vv{x}(k\Delta t) \right]^\intercal \vv{R}^{-1} \left[ \vv{y}_k - \vv{H}\vv{x}(k \Delta t) \right].
\end{align}
Here, $\rho \geq 1$ is an adjustable constant, referred to as the covariance inflation parameter, that discounts past measurements, in part to compensate for model error in $\vv{G}$.
In the situation where the measurement errors ($\eta_k$) are Gaussian and the model $\vv{G}$ is a perfect representation of the dynamical system, i.e., $\vv{G} = \vv{F}$, minimizing Eq.~(\ref{eq:dataAssimilationbase}) with $\rho =1$ will give us the trajectory of the model $\vv{x}(t)$ that is the most likely trajectory of the dynamical system in the Bayesian sense (here, we assume a ``flat" prior likelihood; the choice of prior distribution becomes increasingly unimportant as $T$ increases.)

In forecast applications, the purpose of DA is to determine an estimate $\vv{x}(0)$ of the current system state, which is used as an initial condition for Eq.~(\ref{eq:imperfect}) to forecast $\vv{x}(t)$ for $t > 0$.  As such, the trajectory $\vv{x}(t)$ for $t<0$ that minimizes the cost function $J_0$ is often not computed explicitly; only its final state $\vv{x}(0)$ is needed as the algorithmic output.  DA is typically done sequentially, using the state estimate from an earlier time to compute the estimate for the current time. To be more precise, consider the cost function $J_j$ that generalizes Eq.~(\ref{eq:dataAssimilationbase}) as follows: 
\begin{align}\label{eq:dataAssimilation}
J_j(\vv{x}(t)) = \sum_{k = -T}^j \rho^{k-j} \left[\vv{y}_k - \vv{H}\vv{x}(k\Delta t) \right]^\intercal \vv{R}^{-1} \left[ \vv{y}_k - \vv{H}\vv{x}(k \Delta t) \right].
\end{align}
Then, 
\begin{align}\label{eq:dataAssimilationupdate}
    J_j(\vv{x}(t)) = \rho^{-1}J_{j-1}(\vv{x}(t)) + \left[\vv{y}_j - \vv{H}\vv{x}(j\Delta t) \right]^\intercal \vv{R}^{-1} \left[ \vv{y}_j - \vv{H}\vv{x}(j \Delta t) \right].
\end{align}
Suppose for some $-T < j <= 0$ that $\vv{x}^a_{j-1}$ represents an estimate of $\tilde{\vv{x}}((j-1)\Delta t)$ based on the cost function $J_{j-1}$; i.e., based on measurements $\vv{y}_k$ up to and including $k=j-1$. The next estimate $\vv{x}_j^a$ is computed in two steps as follows:
\begin{enumerate}
    \item \textbf{Forecast}: The model, Eq.~(\ref{eq:imperfect}), is applied from time $(j-1)\Delta t$ to $j\Delta t$, using $\vv{x}^a_{j-1}$ as the initial condition, to obtain an estimate $\vv{x}^b_j$ to $\tilde{\vv{x}}(j\Delta t)$, called the ``background'' state at time $j\Delta t$.
    \item \textbf{Analysis}: The background $\vv{x}^b_j$ is used along with the current measurement vector $\vv{y}_j$ to compute the ``analysis'' state $\vv{x}^a_j$ that estimates the minimizer of $J_j$ at time $j\Delta t$.
\end{enumerate}
These two steps are repeated to compute $\vv{x}_{j+1}^b$, $\vv{x}_{j+1}^a$, and so on.

Different DA methods perform the analysis step differently.  In some methods, including those based on the Kalman filter, the cost functions are approximated by quadratic functions of $\vv{x}(j\Delta t)$, so that the right side of Eq.~(\ref{eq:dataAssimilationupdate}) becomes

\begin{align}\label{eq:dataAssimilationupdateKalman}
    \rho^{-1}\left[\mathbf{x}(j\Delta t)-\mathbf{x}_{j}^{b}\right]^\intercal\left(\mathbf{P}_{j}^{b}\right)^{-1}\left[\mathbf{x}(j\Delta t)-\mathbf{x}_{j}^{b}\right] + \left[\vv{y}_j - \vv{H}\vv{x}(j\Delta t) \right]^\intercal \vv{R}^{-1} \left[ \vv{y}_j - \vv{H}\vv{x}(j \Delta t) \right].
\end{align}
Thus, the quantity to be minimized is a sum of (squared) weighted distances from $\vv{x}(j\Delta t)$ to the background state (which is forecasted from previous observations) and to the current observations.  Here $\vv{P}^b_j$ is a covariance matrix associated with the background state at time $j\Delta t$.  In an ensemble Kalman filter [\onlinecite{evensen2003ensemble, houtekamer_data_1998, burgers1998analysis}], $\vv{P}^b_j$ is computed as the sample covariance of an ensemble of forecasts.  The parameter $\rho$ effectively inflates this covariance, compensating for (in addition to model error) sampling errors due to finite ensemble size and the effect of nonlinearity on the error dynamics [\onlinecite{bishop2001adaptive, whitaker2002ensemble}].

In the Appendix, we describe the particular DA method we use, the ``Ensemble Transform Kalman Filter'' (ETKF) [\onlinecite{bishop2001adaptive, wang2004better}], a type of ensemble Kalman filter.  We emphasize that the approach we describe in Sec.~\ref{sec:mletkf-algorithm} can be used with any DA method.  The forecast and analysis steps we have described are applied iteratively to produce a time series of analysis states $\vv{x}^a_j$.  We will use this time series in Sec.~\ref{sec:mletkf-algorithm} to train the ML component of our method.

\subsection{ML-DA Algorithm}\label{sec:mletkf-algorithm}

In what follows, we use a reservoir computer implementation of our ML-DA algorithm based on Refs.~\onlinecite{jaeger2004harnessing, lukovsevivcius2012practical}, which is similar to that of Ref.~\onlinecite{pathak2018hybrid}. The main component of a reservoir computer is the reservoir network. The reservoir network is a sparse, randomly generated graph with weighted, directed edges and dynamical nodes whose states evolve in response to input signals transported along network edges. The network topology can be represented completely by a weighted adjacency matrix, denoted $\vv{A}$. We choose $\vv{A}$ to be a $D_r \times D_r$ sparse, randomly generated matrix. The network is constructed to have an average in-degree (number of incoming edges per node) denoted by $\langle d \rangle$, and the nonzero elements of $\vv{A}$, representing the edge weights in the network, are initially chosen independently from the uniform distribution over the interval $[0,1]$. All the edge weights in the network are then uniformly scaled via multiplication of the adjacency matrix by a constant factor to set the largest magnitude eigenvalue of the matrix to a desired value $\omega$, which is called the ``spectral radius'' of $\vv{A}$. The state of the reservoir, given by the vector $\vv{r}(t)$, consists of the components $r_n$ for $1\leq n \leq D_r$ where $r_n(t)$ denotes the scalar state of the $n^{th}$ node in the network. 
The reservoir is coupled to the $M$ dimensional input through a $D_r \times M$ dimensional matrix $\vv{W}_{in}$. Each row of the matrix $\vv{W}_{in}$ has exactly one nonzero element which is independently and randomly chosen from the uniform distribution on the interval $[-\zeta,\zeta]$. We choose the locations of these nonzero elements such that each node receives input from only one input variable and each input variable is input to an equal number of nodes.

\begin{figure}[ht]
\centering
\includegraphics[width = 0.7\textwidth]{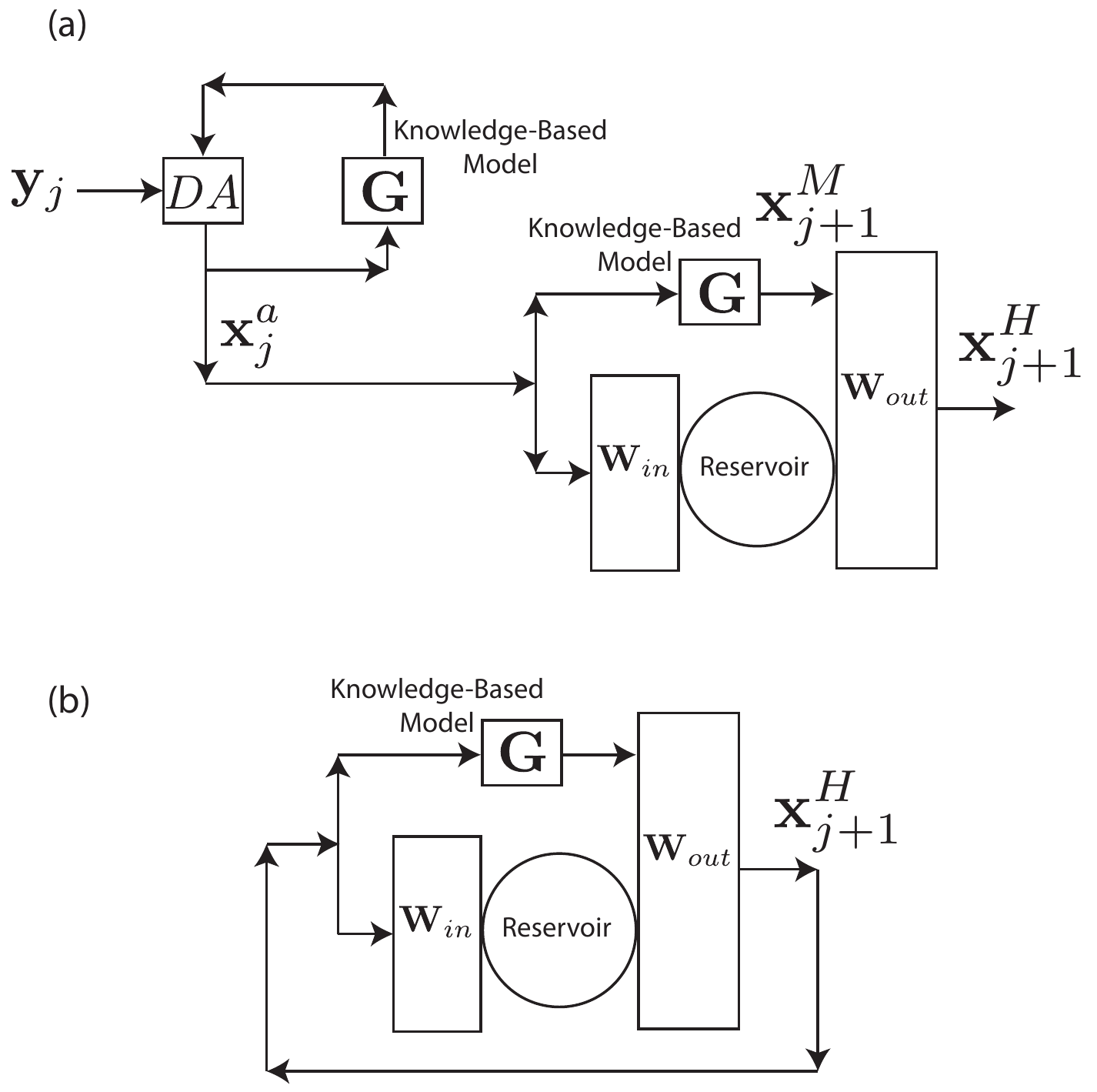}
\caption{(a) Training the ML-DA prediction scheme and (b) forecasting as described in Sec.~\ref{sec:mletkf-algorithm}. The output matrix in (a), $\mathbf{W}_{out}$, is unknown at the time of training. The goal of training is to determine $\mathbf{W}_{out}$ such that $\mathbf{x}^H_j$ best approximates $\mathbf{x}^a_j$ in the regularized least-squares sense.}
\label{fig:mletkf}
\end{figure}

As outlined at the beginning of Sec.~\ref{sec:dynamical}, we assume that the finite-dimensional representation of the true dynamical system is given by Eq.~(\ref{eq:dynamical}). We have $T+T_s$ measurements given by $\lbrace \vv{y}_j \rbrace$, in the interval $-T-T_s \leq j \leq 0$, which are related to the true state of the dynamical system by Eq.~(\ref{eq:measurement}). We refer to $T$ and $T_s$ as the training and synchronization times, where $T\gg T_s$. We further assume that the model $\vv{G}$ (Eq.~(\ref{eq:imperfect})) is imperfect. Using the model $\vv{G}$ and the DA algorithm outlined in Sec.~\ref{sec:nonlinKal}, we can obtain an analysis state $\vv{x}^a_j$ at each time step $j$, $-T-T_s \leq j \leq 0$. We thus create a set of ``analysis'' states $\lbrace \vv{x}_j^a \rbrace_{-T-T_s\leq j \leq 0}$. We are interested in forecasting the state of the dynamical system for $j > 0$. In the standard forecast approach (without ML), one would predict the future state of the dynamical system (Eq.~(\ref{eq:dynamical})) using $\vv{x}^a_0$ as the initial condition and Eq.~(\ref{eq:imperfect}) as the dynamical model. We will call this forecast the baseline forecast, to which we will compare the forecast made by our ML-DA algorithm. The ML-DA algorithm is a natural extension of the algorithm proposed in Ref.~\onlinecite{pathak2018hybrid} for hybrid forecasting of chaotic dynamical systems by using ML in conjunction with a knowledge-based model. Figure~\ref{fig:mletkf} shows a schematic illustration of the ML-DA scheme described below.

\subsubsection{Training}
\label{sec:hybrid-training}
\begin{enumerate}
\item Use the model $\vv{G}$ to create a set of forecasts from the latest $T$ analysis states $\vv{x}^a_j$, $-T \leq j \leq 0$. We denote these time-$\Delta t$ forecasts by $\vv{x}^M_j$, $-T+1 \leq j \leq 1$.
\item Initialize the reservoir state $\vv{r}_{-T-T_s}$ to a random value.
\item Evolve the reservoir computer using the following equation
\begin{align}\label{eq:reservoir-update}
\vv{r}_{j+1} &= \tanh[\vv{A}\vv{r}_j + \vv{W}_{in} \vv{x}^a_j]
\end{align}
for $-T-T_s \leq j \leq 0$. Here, $\tanh$ is applied element-by-element to its input vector.
\item Find a set of output weights $\vv{W}_{out}$ so that 
\begin{align}
\mathbf{x}^H_j = \vv{W}_{out}\left[ \begin{matrix}
\vv{r}_j \\ \vv{x}^M_j
\end{matrix} \right] \simeq \vv{x}^a_j
\end{align}
for $-T+1 \leq j \leq 0$. The matrix $\vv{W}_{out}$ is computed using regularized least squares regression. Thus, we find the $\vv{W}_{out}$ that minimizes the following cost function:
\begin{align}
\ell(\vv{W}_{out}) = \sum_{j=-T+1}^0 \lVert \vv{W}_{out}\left[ \begin{matrix}
\vv{r}_j \\ \vv{x}^M_j
\end{matrix} \right] - \vv{x}^a_j\rVert^2 + \beta\lVert \vv{W}_{out} \rVert^2.
\end{align}
Notice that we have not used the first $T_s$ reservoir states to find the optimal $\vv{W}_{out}$. This is to ensure that the reservoir states used in the computation of $\vv{W}_{out}$ do not depend strongly on the initial random value of the reservoir state, $\mathbf{r}_{-T-T_s}$, and thus are ``synchronized'' to the measurements that drive Eq.~\ref{eq:reservoir-update} through the analysis states $\vv{x}^a_j$.
\end{enumerate}

\subsubsection{Prediction}\label{sec:hybrid-prediction}
We now describe the hybrid forecast model that uses the trained reservoir along with the knowledge-based model $\vv{G}$ to forecast the state $\vv{x}(j\Delta t)$ for $j>0$. 

The hybrid reservoir runs in a closed-loop feedback mode according to the following steps:

\begin{enumerate}
\item Obtain the hybrid ML prediction at time $j\Delta t$ (denoted $\vv{x}^H_j$) according to:
\begin{align}\label{eq:hybrid_pred_a}
\vv{x}^H_j = \vv{W}_{out} \left[ \begin{matrix} \vv{r}_j \\ \vv{x}^M_j \end{matrix} \right].
\end{align}
The reservoir state and knowledge-based forecasts at the first step, $\mathbf{r}_1$ and $\mathbf{x}^M_1$, are computed from the final training analysis $\vv{x}^a_0$ according to steps 1 and 3 of the training procedure.
\item Compute the imperfect model forecast at time $(j+1)\Delta t$, $\vv{x}^M_{j+1}$, by evolving $\vv{x}^H_j$ using the imperfect model $\vv{G}$.
\item Evolve the reservoir according to
\begin{align}\label{eq:hybrid_pred_b}
\vv{r}_{j+1} &= \tanh\left[\vv{A}\vv{r}_j + \vv{W}_{in} \vv{x}_j^H\right]
\end{align}
\item Increment $j$ and repeat steps 1 to 3.
\end{enumerate}
If one wants to initialize a forecast at a time $J\Delta t$ for $J > 0$, one should continue to run the DA procedure for $J$ more steps after the training time period, set $\vv{x}^H_J = \vv{x}^a_J$, then proceed as above starting with step 2 for $j = J$. The $J$
DA cycles after time 0 can be performed using the
trained hybrid model in place of the knowledge-based model, similarly
to the approach described in the next section.

We note that the ensemble-based DA method we use could be used to initialize an ensemble forecast. However, for simplicity, our results use only forecasts from the ensemble mean.

\subsection{Iterated ML-DA Algorithm}\label{sec:iterated-mletkf}
\begin{figure}[ht]
\centering
\includegraphics[width = 0.7\textwidth]{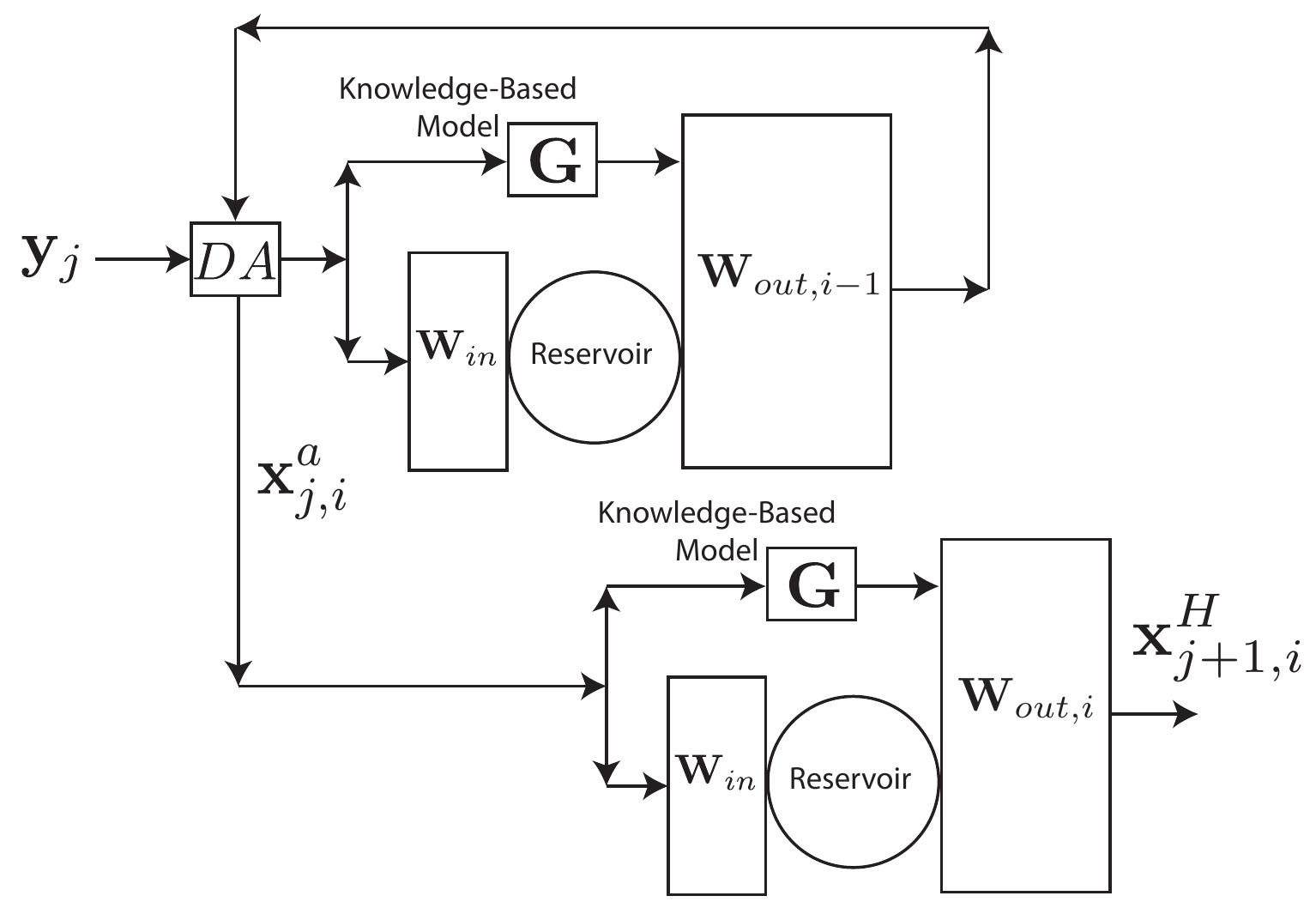}
\caption{Training subsequent iterations ($i\geq 2$) of the ML-DA algorithm using the synchronized hybrid model during DA as described in Sec.~\ref{sec:iterated-mletkf}. The index $i$ denotes the current iteration of the algorithm, and we consider the training procedure shown in Fig.~\ref{fig:mletkf}(a) to occur at $i = 1$. At iteration $i$, $\mathbf{W}_{out,i-1}$ is known from the previous training iteration, while we endeavor to determine $\mathbf{W}_{out,i}$ such that $\mathbf{x}^H_{j,i}$ best approximates $\mathbf{x}^a_{j,i}$ in the regularized least-squares sense.}
\label{fig:iterated-mletkf}
\end{figure}

Once we have trained a hybrid model using the process described in Sec.~\ref{sec:hybrid-training}, we can iterate the training procedure in the following sense. An iterative approach, initialized differently, is described in Ref. \onlinecite{brajard2020combining}. As shown in Fig.~\ref{fig:iterated-mletkf}, we replace the knowledge-based model used during DA with the previously trained hybrid model. At iteration $i=1$, the hybrid is trained using the analysis states produced using only the knowledge-based model. We denote the output matrix to be determined by the training, the state of the reservoir at time $j\Delta t$, the analysis state at time $j\Delta t$, and the hybrid model predictions for time $j\Delta t$ at iteration $i$ as $\mathbf{W}_{out,i}$, $\mathbf{r}_{j,i}$, $\mathbf{x}^a_{j,i}$, and $\mathbf{x}^H_{j,i}$, respectively.

Training proceeds as in Sec.~\ref{sec:hybrid-training} with the following modifications.
\subsubsection{Training}
\begin{enumerate}
    \item For $-T-T_s \leq j < -T$, we use the knowledge-based model to perform DA, producing analysis states $\mathbf{x}^a_{j,i}$. We do not use the hybrid model during this synchronization time period because the reservoir states $\vv{r}_{j,i}$ might not yet be synchronized with the measurements. These reservoir states are generated according to step 3 of Sec.~\ref{sec:hybrid-training}, i.e.,
    \begin{align}\label{eq:evolve-res-iter}
        \mathbf{r}_{j+1,i} = \tanh[\mathbf{A}\mathbf{r}_{j,i} + \mathbf{W}_{in}\mathbf{x}^a_{j,i}].
    \end{align}
    \item For $T\leq j \leq 0$, we use the hybrid model to perform DA in place of the knowledge-based model, producing analysis states $\mathbf{x}^a_{j,i}$, as depicted in Fig.~\ref{fig:iterated-mletkf}. We continue to use Eq.~(\ref{eq:evolve-res-iter}) to evolve the reservoir states. 
    \item Using these new analysis states, we follow the steps contained in Sec.~\ref{sec:hybrid-training} to compute a new hybrid output matrix, $\mathbf{W}_{out,i+1}$.
    \item Once the new hybrid output matrix has been computed, predictions can be produced by following Sec.~\ref{sec:hybrid-prediction}.
\end{enumerate}
In the case of an ensemble-based assimilation method (such as the ETKF described in Appendix~\ref{sec:nonlinKal}), we augment each ensemble member to include, in addition to the system state $\mathbf{x}^{a,k}_j$, a corresponding reservoir state $\mathbf{r}^k_j$. This reservoir state is synchronized to the corresponding ensemble system state by evolving it as in Eq.~(\ref{eq:evolve-res-iter}):
\begin{align}
    \mathbf{r}^k_{j+1,i} = \tanh[\mathbf{A}\mathbf{r}^k_{j,i} + \mathbf{W}_{in}\mathbf{x}^{a,k}_{j,i}].
\end{align}
Then, $\vv{r}^k_{j+1,i}$ is used in the hybrid model to produce the $k^{\textrm{th}}$ background ensemble member for the next analysis cycle. We note that the reservoir state(s) at $j=-T+1$ needed to begin DA using the hybrid model must only be computed at the first iteration. Afterwards, these initial synchronized states may be reused at the beginning of subsequent DA cycles.
\section{Results}\label{sec:results}

We demonstrate the forecast performance of the ML-DA algorithm in comparison with the performance of baseline for forecasts prepared without using MD. We consider two dynamical systems as our examples: the Lorenz `63 system~[\onlinecite{lorenz1963deterministic}] and the Kuramoto-Sivashinsky (KS) system~[\onlinecite{kuramoto1976persistent,sivashinsky1977nonlinear}]. For the latter system, we additionally demonstrate the decrease in error in the analysis states from using the iterative method described in Sec.~\ref{sec:iterated-mletkf}.
\subsection{Experimental Design}
For the purpose of evaluating forecast performance, simulated true states $\hat{\vv{x}}_j$ are generated by the perfect model dynamics in the time interval $-T-T_S \leq j \leq P$, where $T$ and $T_S$ are described in Sec.~\ref{sec:mletkf-algorithm}. The data in the interval $1 \leq j \leq P$ is set aside to verify the baseline and ML-DA forecasts but is not otherwise used, since it is considered to be in the future. We assume that we only have access to the measurements $\lbrace \vv{y}_j \rbrace_{-T-T_S \leq j \leq 0 }$ generated from the simulated true states via Eq.~(\ref{eq:measurement}) and do not know the perfect model, Eq.~(\ref{eq:dynamical}). We further assume that we have access to an imperfect model $\vv{G}$ (Eq.~\ref{eq:imperfect}). We use this imperfect model to perform DA, to train the ML-DA, and to prepare both the baseline and ML-DA forecasts for $j > 0$.

For DA, we use the ETKF, as described in the Appendix, with 15 ensemble members for the Lorenz '63 system and 30 ensemble members for the KS system.  To initialize the forecasts and the ML training, we perform DA using the measurements $\vv{y}_j$ ($-T-T_S \leq j \leq 0$) and the model $\vv{G}$ to obtain the analysis states $\vv{x}^a_{-T-T_s}, \dots, \vv{x}^a_{0}$.

\subsubsection{Baseline Forecasts: } Using $\vv{x}^a_0$ as the initial condition, we compute a forecast using the model $\vv{G}$ from time $j = 1$ to $j = P$. We call this forecast $\vv{x}^{f, base}_j$, $1 \leq j \leq P$.
\subsubsection{ML-DA Forecasts: } We follow the ML-DA scheme detailed in Sec.~\ref{sec:mletkf-algorithm} using a reservoir computer with the parameters listed in Table~\ref{tab:hyperparameters} for the Lorenz `63 and the Kuramoto-Sivashinsky (KS) dynamical systems. We use the measurements $\vv{y}_j$ ($-T-T_S \leq j \leq 0$) and the imperfect model $\vv{G}$ corresponding to the Lorenz and KS systems respectively to train the reservoir computer. The exact form of the measurements and of the imperfect model is described in Secs.~\ref{sec:lorenzres} and \ref{sec:ksres}. We then forecast using the ML-DA scheme from $j = 1$ to $j = P$. We call this forecast $\vv{x}^{f, ml}_j$, $1 \leq j \leq P$.

\begin{table}
\centering
\begin{tabular}{c|c|c}
Hyperparameter & Lorenz 63 & KS\\
\hline
$D_r$ & $1000$ & $2000$\\
$\langle d \rangle$ & 3 & 3\\
$\omega$ & 0.9 & 0.6 \\
$\zeta$ & 0.1 & 1.0 \\
\end{tabular}
\caption{Reservoir Hyperparamenters}
\label{tab:hyperparameters}
\end{table}
\subsubsection{Normalized RMS error}
The normalized RMS error in the baseline forecast ($e_j^{base}$) and the RMS error in ML-DA forecast ($e^{ml}_j$) are calculated as follows:
\begin{align}\label{eq:rms-error}
e_j &= \frac{\lVert \vv{x}^f_j - \hat{\vv{x}}_j \rVert}{\sqrt{\langle \lVert \hat{\vv{x}}_j \rVert^2 \rangle}},  \hspace{0.5in} 1 \leq j \leq P,
\end{align}
where $||\dots||$ denotes the Euclidean norm, $\langle\dots\rangle$ denotes an average over the prediction interval ($1\leq j \leq P$), and $e_j$, $\vv{x}^f_j$ denote either $e^{base}_j$, $\vv{x}^{f,base}_j$ or $e^{ml}_j$, $\vv{x}^{f,ml}_j$. The superscript $f$ in $\vv{x}^f_j$ stands for forecast.

\subsubsection{Valid Time}\label{sec:valid-time}
We define the Valid Time (VT) as the time $j \Delta t$ at which the RMS error ($e_j$) first exceeds a threshold $\kappa$ chosen to be $0.9$. In order to report the forecast quality on a meaningful natural time scale, we normalize the valid time by the `Lyapunov time' of the dynamical system. The Lyapunov time is defined as the inverse of the Largest Lyapunov Exponent (LLE), denoted $\Lambda_{max}$, of the dynamical system being forecast. In a chaotic system, infinitesimal errors grow exponentially on average as $\exp(\Lambda_{max}t)$. Thus, $\Lambda_{max}^{-1}$ is a natural time scale for evaluating forecast quality in a chaotic system.

Since forecast Valid Time can vary considerably depending on what part of the attractor forecasts are initialized from, we present Valid Time results as box plots reflecting 100 different runs for the Lorenz System and 50 runs for the KS system. Each run uses a different initial condition for the training data (and therefore for the following forecast) and a different random realization of the reservoir. All box plots depict the median, $25^{\textrm{th}}$ and $75^{\textrm{th}}$ percentiles, and the $5^{\textrm{th}}$ and $95^{\textrm{th}}$ percentiles (depicted by dashed lines).

\subsubsection{Analysis RMS Error}
For the iterative method in Sec.~\ref{sec:iterated-mletkf}, we evaluate the error in the analysis state at a particular iteration $i$ by computing the normalized RMS error over the entire analysis period,
\begin{align}
    \label{eq:analysis-error}
    e^a_{tot,i} = \sqrt{\frac{\langle \lVert \mathbf{x}^a_{j,i}-\mathbf{x}_j \rVert^2\rangle}{\langle \lVert \mathbf{x}_j \rVert ^2\rangle}},
\end{align}
where $\langle \dots \rangle$ denotes an average over the DA interval ($-T-T_S \leq j\Delta t \leq 0$). In addition, we evaluate the analysis error in the measured or unmeasured system variables by including only the particular set of variables in the norms in Eq.~(\ref{eq:analysis-error}).

\subsection{Lorenz 63}\label{sec:lorenzres}

The Lorenz system is described by the equations,

\begin{align}\label{eq:lorenz1}
\frac{dX_1}{dt} &= -aX_1 + aX_2\\\label{eq:lorenz2}
\frac{dX_2}{dt} &= bX_1 - X_2 - X_1X_3\\\label{eq:lorenz3}
\frac{dX_3}{dt} &= -cX_3 + X_1X_2
\end{align}
where $a = 10$, $b = 28$, and $c = 8/3$. In this example, the true dynamical system, Eq.~(\ref{eq:dynamical}), is given by Eqs.~(\ref{eq:lorenz1}-\ref{eq:lorenz3}). We obtain simulated data by integrating Eqs.~(\ref{eq:lorenz1}-\ref{eq:lorenz3}) using a fourth order Runge-Kutta solver with time step $\tau = 0.01$. We sample the time series at intervals $\Delta t = 0.01$. This data is taken to represent the true system state $\hat{\vv{x}}(j\Delta t)$, $-T-T_S \leq j \leq P$. We use the data in the interval $-T-T_S \leq j \leq 1$ to create simulated measurements 
\begin{align}
\vv{y}_j = \vv{H} \hat{\vv{x}}_j + \boldsymbol{\eta}_j.
\end{align}
We will study two examples with two different forms of the measurement operator $\vv{H}$, one in which only the $X_1$ variable is measured,
\begin{align}
    \vv{H} = \begin{bmatrix} 1 & 0 & 0
    \end{bmatrix},
\end{align}
and another in which the $X_1$ and $X_3$ variables are measured,
\begin{align}
    \vv{H} = \begin{bmatrix} 1 & 0 & 0\\0 & 0 & 1
    \end{bmatrix}.
\end{align}
Measurement noise $\boldsymbol{\eta}_j$ with a Gaussian distribution that has mean zero and covariance matrix $\vv{R}$ is added to the measurements. The covariance matrix is diagonal, with the diagonal elements equal to $\sigma_{noise}^2$, which corresponds to the situation in which the errors in the measurements of the different state vector components are statistically independent. In Sec.~\ref{sec:lorenz-rho} and Sec.~\ref{sec:lorenz-epsilon}, we let $\sigma_{noise} = 0.1$. For comparison, the standard deviations $\sigma_{X_1}, \sigma_{X_2}, \sigma_{X_3}$ of the variables $X_1, X_2, X_3$ are $7.9$, $8.9$ and $8.6$, respectively. In Sec.~\ref{sec:lorenz-sigma}, we consider the effect of varying $\sigma_{noise}$.

We let the imperfect model equations be given by

\begin{align}\label{eq:lorenzImperfect}
\frac{dX_1}{dt} &= -aX_1 + aX_2,\\
\frac{dX_2}{dt} &= b(1+\epsilon)X_1 - X_2 - X_1X_3,\\
\frac{dX_3}{dt} &= -cX_3 + X_1X_2.
\end{align}
Thus, the imperfect model $\vv{G}$ differs from the perfect model $\vv{F}$ in the parameter $b$ by a multiplicative factor $(1+\epsilon)$. Notice also that the model error is in the evaluation of $X_2$, which is unmeasured in our experiments.

\subsubsection{Optimizing the Covariance Inflation}\label{sec:lorenz-rho}

\begin{figure}
\centering
\includegraphics[width = 0.5\textwidth]{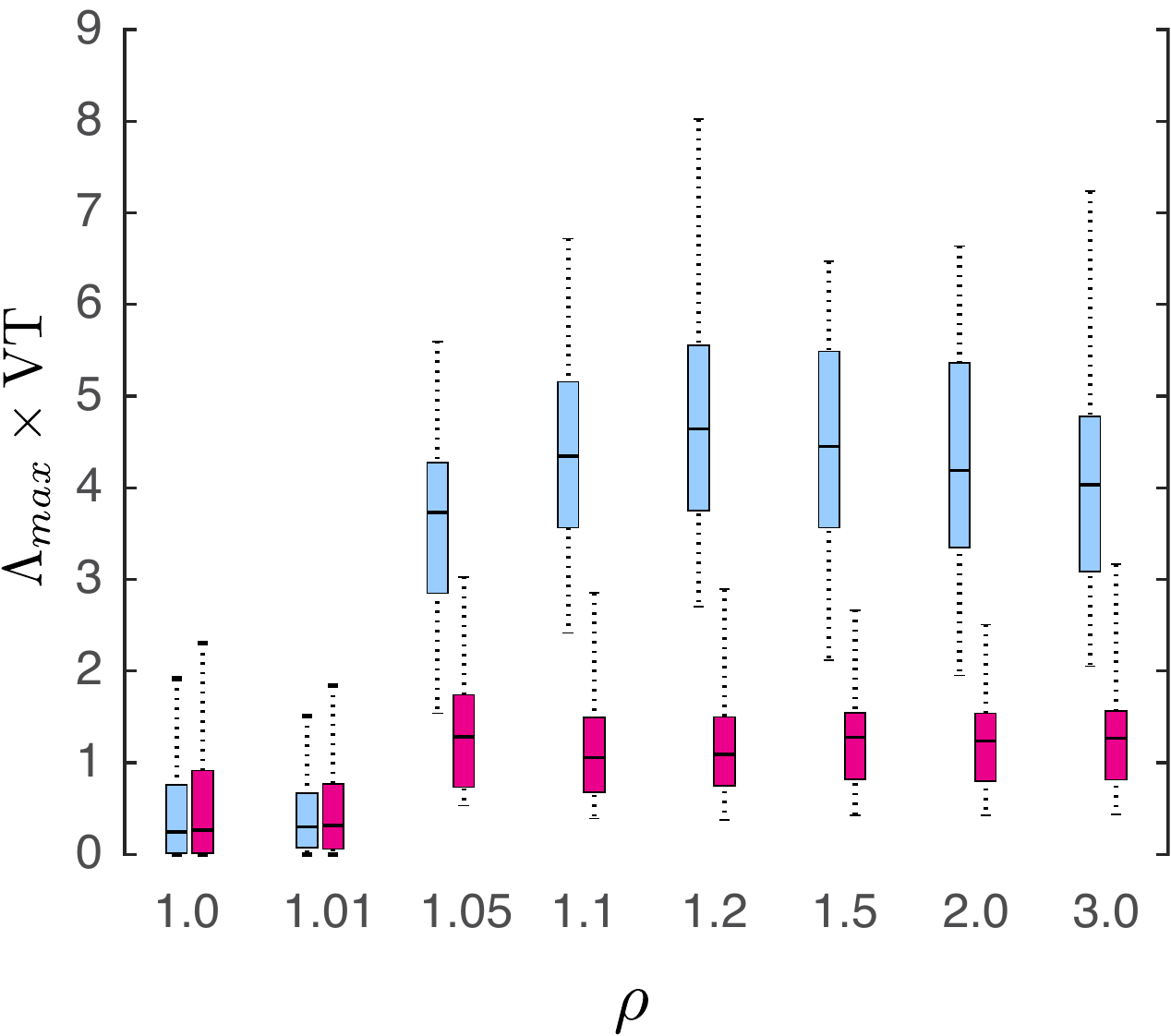}
\caption{Dependence of the Forecast Valid Time (VT) for the Lorenz '63 model on the covariance inflation factor ($\rho$) when $\vv{H} = \begin{bmatrix} 1 & 0 & 0\end{bmatrix}$, measurement noise has magnitude $\sigma_{noise} = 0.1$, and the model error is $\epsilon = 0.1$. The ML-DA forecasts (light blue box plots) outperform the baseline  forecasts (magenta box plots) significantly.}
\label{fig:lorenzRhoPlot}
\end{figure}

The quality of both the baseline and ML-DA forecasts depends on the covariance inflation parameter $\rho$ (see Eqs.~(\ref{eq:dataAssimilation}) and~(\ref{eq:ETKF}).)This parameter is used in the DA cycle that initializes both forecasts, as well as forming the training data for the ML-DA. It is thus crucial that the covariance inflation parameter is optimized independently for both the forecast schemes. In Fig.~\ref{fig:lorenzRhoPlot}, for $\vv{H} = \begin{bmatrix} 1 & 0 & 0\end{bmatrix}$, we demonstrate the dependence of the forecast Valid Time on the covariance inflation parameter $\rho$ when the model error is $\epsilon = 0.1$. Each Valid Time is shown for a set of 100 forecasts (see Sec.~\ref{sec:valid-time}) made with the baseline scheme (magenta box plots) and the ML-DA scheme (light blue box plots). Figure~\ref{fig:lorenzRhoPlot} shows that the ML-DA scheme dramatically improves forecast valid time when the model has substantial error. In particular, the best median VT for the ML-DA (which occurs at $\rho=1.2$) exceeds the best median VT baseline (which occurs at $\rho=1.05$) by about a factor of 3. We speculate that the optimal value of $\rho$ is larger for ML-DA than the baseline for the following reason. Larger values of $\rho$ cause the measurements to be weighted more heavily relative to the background forecast from the imperfect model in the DA procedure. This can be advantageous for the training, which seeks to use the measurements to correct the model error.

\begin{figure}
\centering
\includegraphics[width = 0.7\textwidth]{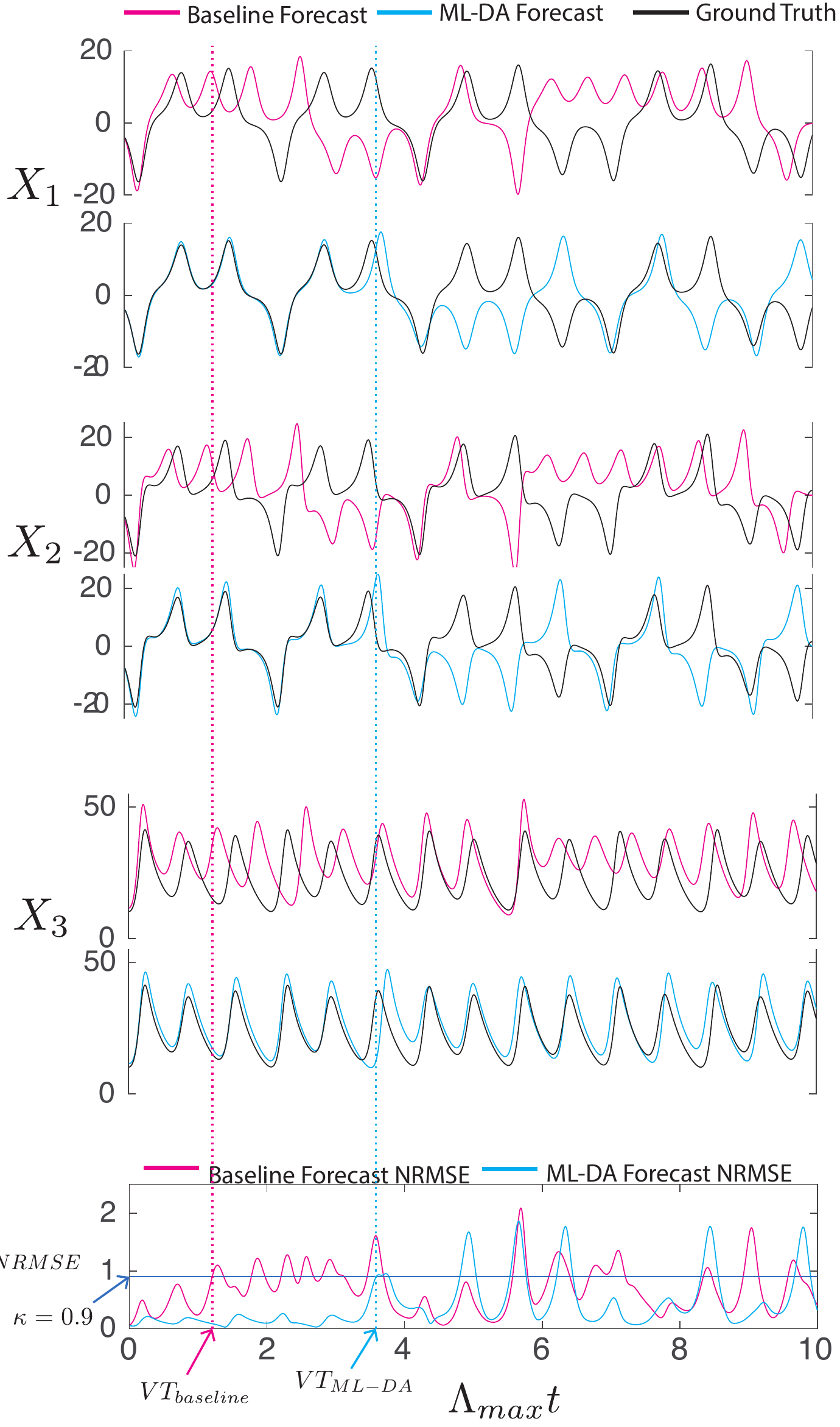}
\caption{Forecast time series plots for the Lorenz '63 model using the ML-DA and baseline scheme for $\vv{H} = \begin{bmatrix} 1 & 0 & 0\end{bmatrix}$ and model error $\epsilon = 0.1$.}
\label{fig:lorenz_plot}
\end{figure}

In the results that follow, for each choice of method,
model error, and noise magnitude, we choose the covariance inflation
$\rho$ to be the value that yields the highest median Valid Time,
among the values shown in Fig.~\ref{fig:lorenzRhoPlot}. For our usual values $\epsilon = 0.1$ and $\rho_{noise} = 0.1$, Figure~\ref{fig:lorenz_plot} shows a typical example comparing the true and predicted time series; notice that the ML-DA scheme improves the forecast for all three variables, including those those that are unobserved.

\subsubsection{Dependence on Model Error}\label{sec:lorenz-epsilon}
\begin{figure}
\centering
\includegraphics[width = 0.5\textwidth]{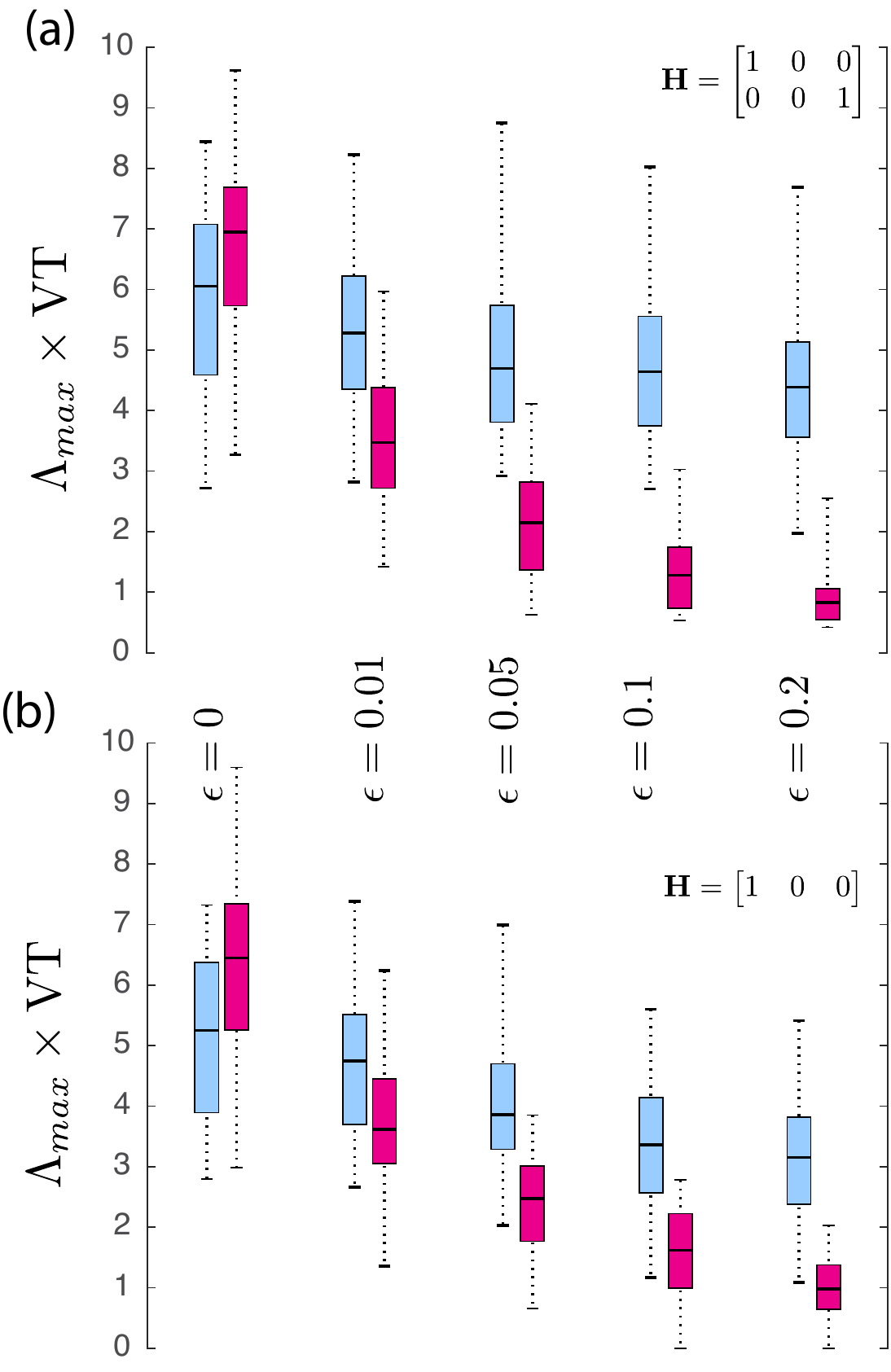}
\caption{Dependence of the Forecast Valid Time for the Lorenz '63 model on the model error ($\epsilon$). The box plots represent the result of 100 independent trials comparing the baseline and the ML-DA forecasts, each using a different Lorenz '63 dataset starting from a random initial condition. The covariance inflation parameter is independently optimized for both sets of forecasts at each value of the model error ($\epsilon$). (a) and (b) differ in the measurement operator used, which is indicated in the figures.}
\label{fig:lorenz_error_dependence}
\end{figure}

Figure~\ref{fig:lorenz_error_dependence} demonstrates the effectiveness of the ML-DA algorithm at different values of the model error ($\epsilon$). As expected, we find that the ML-DA optimized over $\rho$ yields the most improvement relative to the baseline optimized over $\rho$ when the model error is large. In the cast when the model error is zero, the ML-DA algorithm degrades the forecasts somewhat; since there is no model error to correct, the modification made to the model by the ML is spurious.

\subsubsection{Effects of measurement Noise}\label{sec:lorenz-sigma}
\begin{figure}[ht]
\centering
\includegraphics[width = 0.5\textwidth]{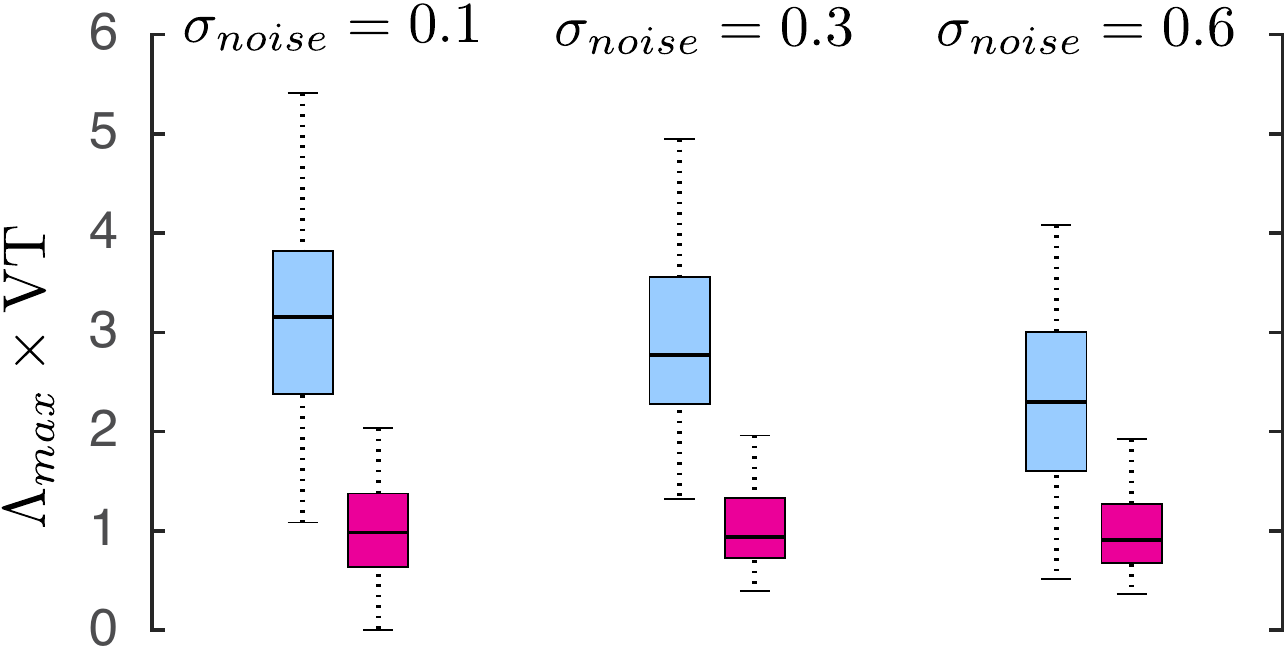}
\caption{Effect of varying amounts of measurement noise on the baseline (magenta box plots) and ML-DA (light blue box plots) forecast valid time for the Lorenz '63 system, with model error $\epsilon = 0.1$. The measurement operator is $\vv{H} = \left[\begin{matrix}
1 & 0 & 0
\end{matrix}\right]$, and simulated uncorrelated Gaussian noise with standard deviation $\sigma_{noise}$ is added to the measurements. For comparison, the standard deviation of the $X_1$ variable is $\sigma_{X_1} \simeq 7.9 $. }
\label{fig:lorenz-noise}
\end{figure}

Figure~\ref{fig:lorenz-noise} shows the effects of varying amounts of measurement noise on the baseline (magenta box plots) and ML-DA (light blue box plots) forecast valid time for the Lorenz system. The measurement operator is $\vv{H} = \begin{bmatrix} 1 & 0 & 0\end{bmatrix}$ and simulated uncorrelated Gaussian noise with standard deviation $\sigma_{noise}$ is added to the measurements. For comparison, the standard deviation of the $X_1$ variable is $\sigma_{X_1} \simeq 7.9 $. We see that the ML-DA outperforms the baseline forecast by a larger amount when the noise is smaller. We note that smaller amounts of noise allow the measurements to be weighted more relative to the background forecast in the DA procedure, yielding training data more suitable to correcting forecast errors.

\subsection{Kuramoto-Sivashinsky (KS) system}\label{sec:ksres}

In this section we consider an example of a spatiotemporal chaotic dynamical system called the the Kuramoto-Sivashinsky (KS) system defined by the Partial Differential Equation (PDE),

\begin{align}\label{eq:ks_exact}
\frac{\partial u(x,t)}{\partial t} + u\frac{\partial u(x,t)}{\partial x} + \frac{\partial^2 u(x,t)}{\partial x^2}  +\frac{\partial^4 u(x,t)}{\partial x^4} &= 0.
\end{align} 

Equation~(\ref{eq:ks_exact}) defines the evolution of a one-dimensional spatiotemporal scalar field $u(x,t)$ on the spatial domain $x \in [0, L)$. We assume periodic boundary conditions so that $u(x+L,t) = u(x,t)$. In this example, we let Eq.~(\ref{eq:ks_exact}) represent the true, unknown space and time continuous dynamics. We obtain a time series of finitite-dimensional representations of the state that we assume to satisfy Eq.~(\ref{eq:dynamical}) by discretizing the domain $[0,L)$ into $Q$ grid-points and integrating Eq.~(\ref{eq:ks_exact}) using a pseudo-spectral PDE solver~[\onlinecite{kassam2005fourth}] with time step $\tau = 0.25$. We sample the time series at intervals of $\Delta t = 0.25$. Thus, our simulated true system states takes the form of  a $Q$-dimensional vector $\hat{\vv{x}}(j\Delta t)$, $-T-T_S \leq j \leq P$. As in the previous example of the Lorenz `63 dynamical system (Sec.~\ref{sec:lorenzres}), we generate simulated measurements by adding random Gaussian noise to the simulated true states. We use the simulated measurements in the interval $-T-T_S \leq j \leq 0$ as the training data for the ML-DA scheme and set aside the data in the interval $1 \leq j \leq P$ for forecast verification. 

We assume that we have access to an imperfect model, $\vv{G}$, of the KS dynamical system given by the equations
\begin{align}
\frac{\partial u(x,t)}{\partial t} + u\frac{\partial u(x,t)}{\partial x} + (1 + \epsilon)\frac{\partial^2 u(x,t)}{\partial x^2}  +\frac{\partial^4 u(x,t)}{\partial x^4} &= 0.
\end{align} 
Thus, if $\epsilon = 0$, then the model represents the true dynamics perfectly, while a nonzero value of $\epsilon$ indicates an imperfect model. We also assume that our measurements are of the form given by Eq.~(\ref{eq:measurement}). The measurement operator $\vv{H}$ is a projection operator that outputs $\Theta$ uniformly spaced measurements (of the $Q$ total variables) in $\hat{\vv{x}}_j$. The measurement noise is Gaussian with zero mean and a diagonal covariance matrix with diagonal elements equal to $\sigma_{noise}^2$, where $\sigma_{noise} = 0.1$. For comparison, the standard deviation of the KS time series variable is $\sigma_{KS} = 1.3$.
\label{sec:ks-epsilon}
\subsubsection{Forecast Quality and Dependence on Model Error}\label{sec:KSforecast}
\begin{figure}[ht]
\centering
\includegraphics[width = 0.7\textwidth]{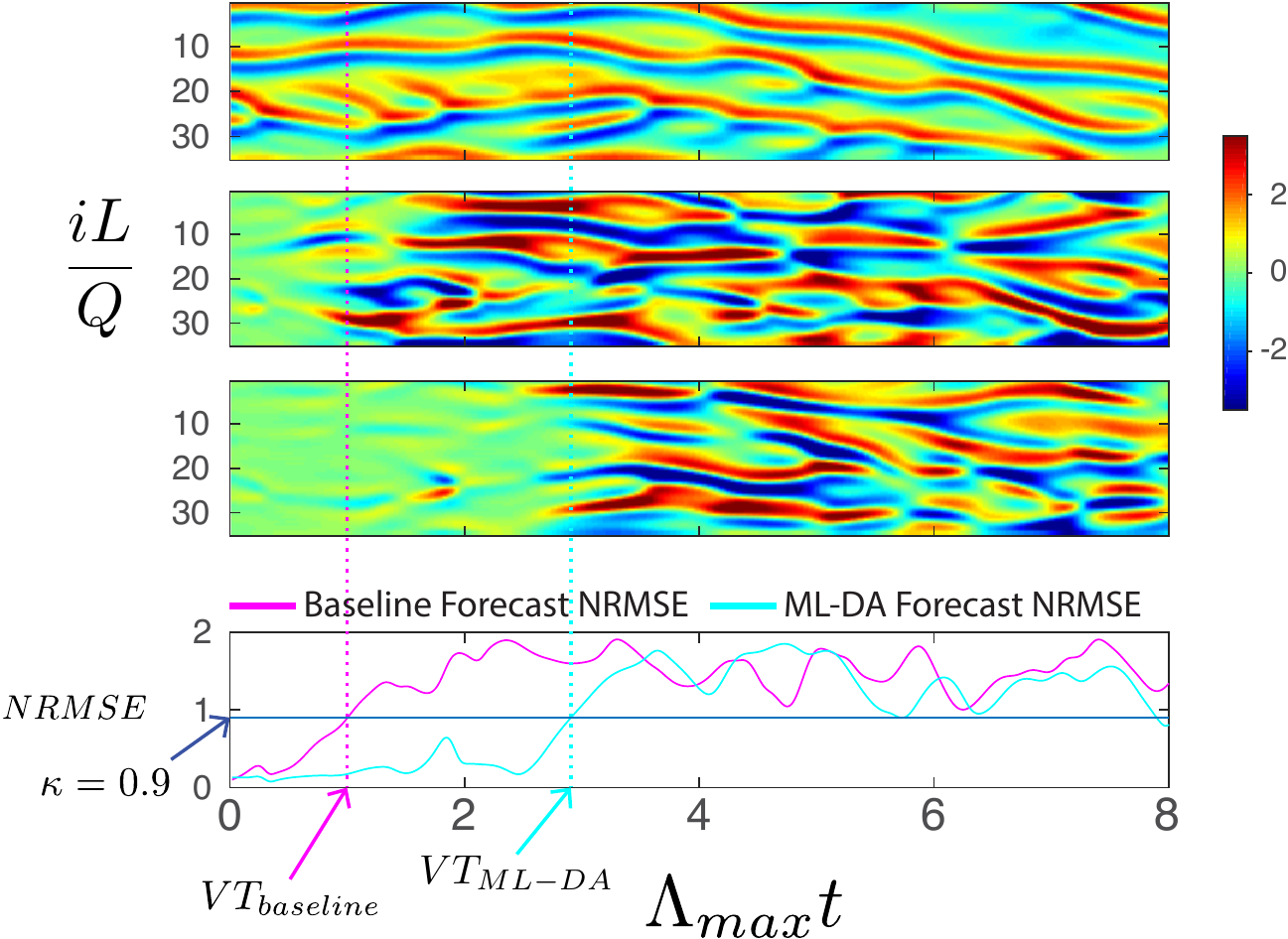}
\caption{The top 3 panels in this figure are spatiotemporal plots with the 64 spatial grid point locations plotted vertically and the Lyapunov time plotted horizontally. From top to bottom in these panels, the color axis shows the true value of $\hat{\vv x}_i(t), i = 1,2, \dots, Q$ from the Kuramoto-Sivashinky system, the difference between the true $\hat{\vv x}_i(t), i = 1,2, \dots, Q$ and the prediction from the baseline scheme, and  the difference between the true $\hat{\vv x}_i(t), i = 1,2, \dots, Q$ and the prediction from the ML-DA scheme. The bottom plot shows the normalized RMS error as function of time for the baseline (red) and ML-DA (blue) schemes, while the dotted vertical black lines mark the valid time of prediction using each scheme. For both schemes, the model error was $\epsilon = 0.1$, and the state was measured at $\Theta = 16$ equally spaced locations of the full $Q=64$ grid points with measurement noise magnitude $\sigma_{noise} = 0.1$.}
\label{fig:stplot}
\end{figure}

To demonstrate the effectiveness of the ML-DA technique, we consider a KS system with $L = 35$, $Q = 64$ and a measurement operator $\vv{H}$ that corresponds to $\Theta = 16$ uniformly spaced measurements. The upper three panels of Fig.~\ref{fig:stplot} shows (from top  to bottom) the time dependence of the true discretized Kuramoto-Sivashinksy equation dynamics, the error in the baseline prediction, and the error in the ML-DA prediction when the model error is $\epsilon = 0.1$. The bottom panel in Fig.~\ref{fig:stplot} shows the normalized RMS error of both predictions. Notice that the ML-DA again significantly outperforms the baseline scheme. 

\begin{figure}[ht]
\centering
\includegraphics[width = 0.5\textwidth]{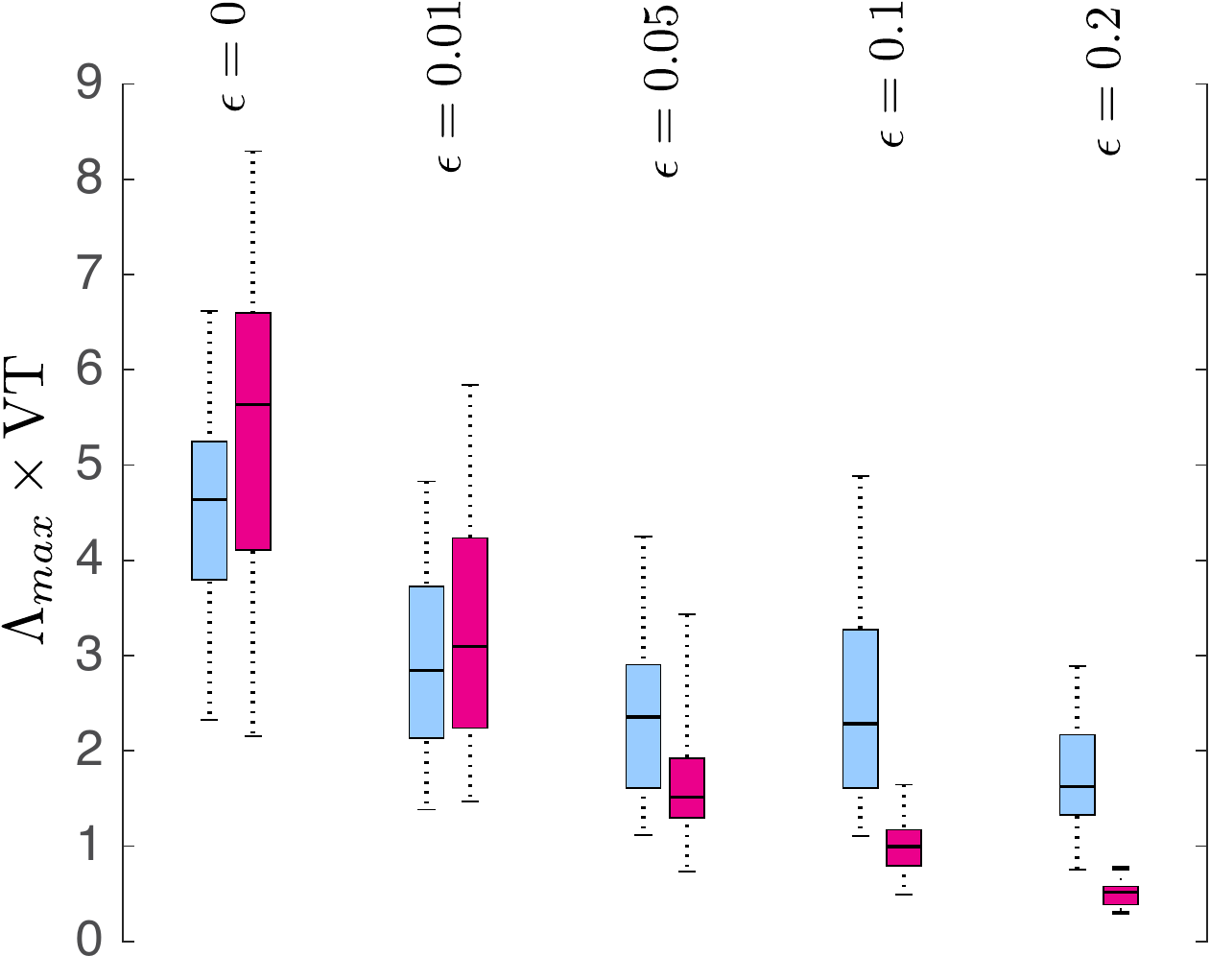}
\caption{Forecast Valid Time of the ML-DA (light blue box plots) and baseline (magenta box plots) schemes at different values of the model error ($\epsilon$).}
\label{fig:ks_error_dependence}
\end{figure}

Figure~\ref{fig:ks_error_dependence} shows the forecast Valid Time for $50$ different forecasts per box plot made with the ML-DA and baseline scheme at four different values of the model error $\epsilon$. As for the Lorenz system, the ML-DA scheme outperforms the baseline scheme when the error in the imperfect model is high, but cannot improve on the baseline when the model error is zero. For this system, the ML-DA also does not improve the forecast Valid Time for $\epsilon = 0.01$, as it is likely to be the case in any system for sufficiently small model error.

\subsubsection{Analysis Error using Iterative ML-ETKF}
\begin{figure}[ht]
\centering
\includegraphics[width = 0.65\textwidth]{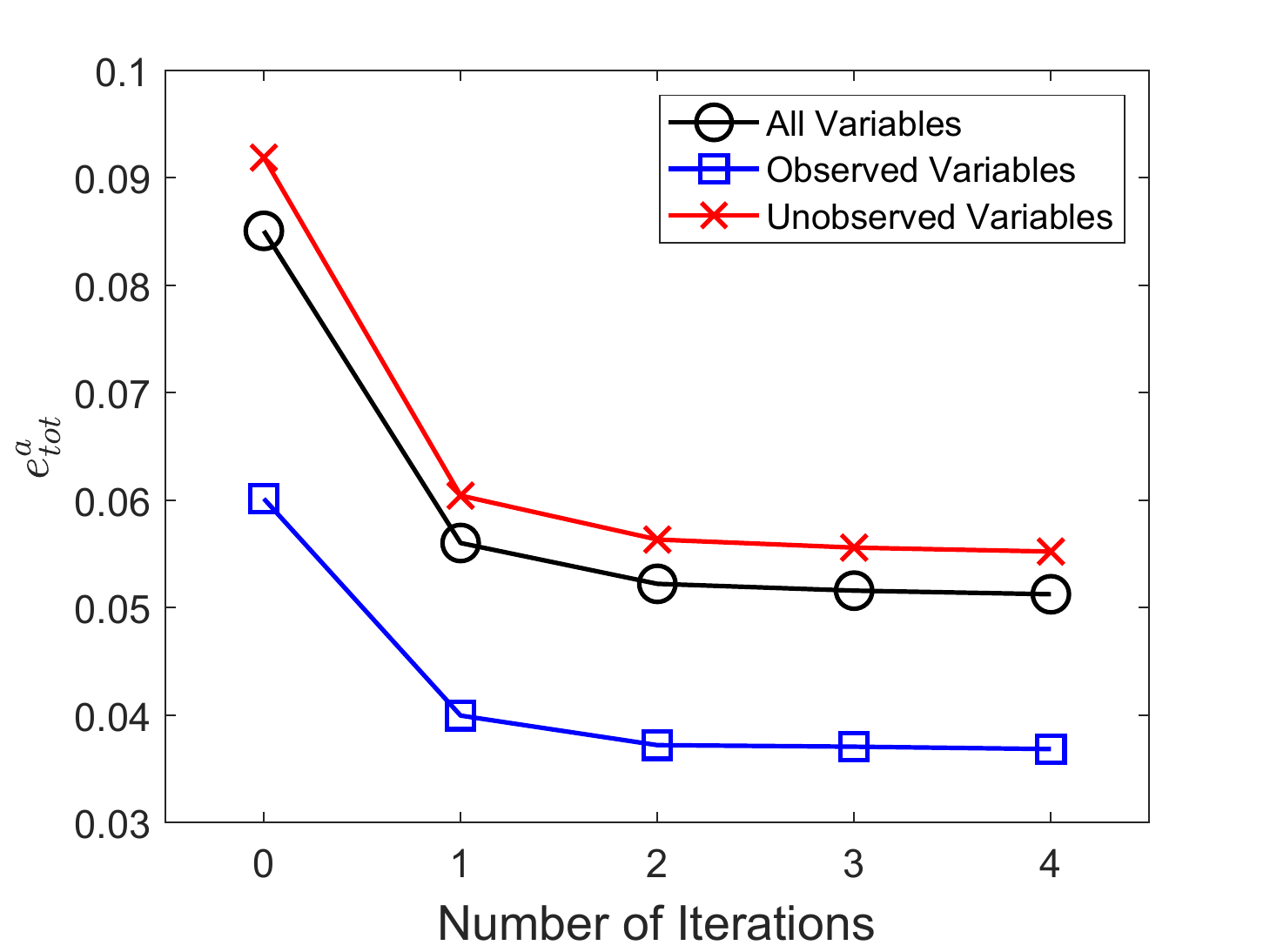}
\caption{Normalized RMS error during DA as we iteratively use the ML-DA to produce analysis time series and use those time series to train the hybrid model. A covariance inflation of $\rho = 2$, which we found to produce the lowest analysis error, was used to produce analysis time series at each step of the iteration. In addition, all knowledge-based models used in this plot had an error of $\epsilon = 0.2$. The plotted values are the median normalized RMS error values computed over 50 different sets of measurement data. The variance in the error over these different data sets is very small; therefore, we have chosen not to plot the standard deviation. At iteration 0, the error values are those from using only the knowledge-based model as our forecast model during DA. We also plot the error values computed over only the measured or the unmeasured variables.}
\label{fig:ks_iterative_analysis_error}
\end{figure}

To demonstrate that the ML-DA hybrid model is also capable of improving the analysis states in a DA cycle, we re-do the DA run that produced the training data using the hybrid model in place of the knowledge-based model. We then re-train the hybrid model, and iterate as described in Sec.~\ref{sec:iterated-mletkf}. Figure~\ref{fig:ks_iterative_analysis_error} shows the error in analysis states computed using iteratively produced hybrid models. Here, iteration $0$ denotes the original set of analysis states computed using only the knowledge-based model. We see that using the first 2 generations of the trained ML-DA to perform DA results in a significant reduction in error in the analyses of both the measured and unmeasured variables. Later iterations give smaller reductions in error. We also measured the forecast Valid Time for each hybrid model trained on the iteratively-produced analysis time series. We found that there was only a small increase in the median Valid Time after one iteration ($\sim 10$\%), and that successive iterations did not improve the Valid Time.
\section{Conclusions and Discussion}
We have demonstrated that DA can successfully extend the hybrid approach introduced in [\onlinecite{pathak2018hybrid}], which trains a machine-learning component based on reservoir computing to improve a knowledge-based forecast model, to the case where a time series of directly-measured model states is not available for training.  Though we use a specific ensemble DA scheme, ETKF, in our numerical experiments, any DA scheme could be used to process the available measurements into estimated model states for use as training data. Using a knowledge-based model rather than a purely data-driven approach allows us to perform DA before training the machine-learning component.  A potential drawback to this approach is that biases present in the knowledge-based model will affect the training data, reducing the effectiveness of the training in correcting for these biases.  Using the iterative approach described in Sec.~\ref{sec:iterated-mletkf}, however, we are able to reduce the effect of these biases on the analysis errors by using the previously trained hybrid model as the forecast model on the DA. In applications like weather forecasting, where multiple knowledge-based models are available, another approach would be to use a different model in the DA procedure than in the hybrid training and prediction procedure.

Another natural extension of the method described in this article would be to integrate the machine-learning training with the data-assimilation cycle in such a way that the training is updated each time new data becomes available.  The use of reservoir computing as the machine-learning component of our hybrid model allows a large number of parameters (the elements of the output weight matrix) to be trained efficiently using linear regression, which could be done sequentially (like DA) through techniques such as recursive least squares.  It could be fruitful to more closely integrate the state estimation and the estimation of the parameters of the hybrid model. The unified Bayesian framework described in [\onlinecite{bocquet_bayesian_2020}] and [\onlinecite{bocquet2020online}] could be applied to a hybrid model, though a successful implementation may require limiting the number of machine-learning parameters.
\section*{Acknowledgement}
This work was supported by DARPA grant HR00111890044. The work of Istvan Szunyogh was also supported by ONR Award N00014-18-2509.
\section*{Appendix}
\label{sec:nonlinKal}
The particular data assimilation algorithm used in this paper is the Ensemble Transform Kalman Filter (ETKF)~[\onlinecite{bishop2001adaptive, wang2004better}], a variant of the Ensemble Kalman Filter (EnKF)~[\onlinecite{evensen2003ensemble, houtekamer_data_1998, burgers1998analysis}]. Our formulation is based on the local
analysis used in the LETKF [\onlinecite{ott2004local, hunt2007efficient, szunyogh2008local}] for spatially extended systems larger than the KS system used in Sec.~\ref{sec:ksres}.

Recall from Sec.~\ref{sec:dataAssimilation} that the analysis state at time $j\Delta t$ is computed from the analysis state at time $(j-1)\Delta t$ by a forecast step followed by an analysis step.  An EnKF uses an ensemble $\lbrace \vv{x}^{a,k}_j\rbrace_{1 \leq k \leq E}$ of analysis states whose mean $\bar{\vv{x}}^a_j$ represents the estimated state and whose covariance $\vv{P}^a_j$ represents the uncertainty in the estimate.  We now describe the forecast and analysis steps for the ensemble.

In an EnKF, the background ensemble members $\vv{x}^{b,k}_j$ are forecast separately using the model, Eq.~(\ref{eq:imperfect}), so that
\begin{align}
\vv{x}^{b,k}_{j} = \vv{x}^k(j\Delta t)\quad \text{where}\quad \frac{d\vv{x}^k}{dt} = \vv{G}[\vv{x}^k(t)],\quad \vv{x}^k((j-1)\Delta t) = \vv{x}^{a,k}_{j-1}.
\end{align}
The sample covariance $\vv{P}_j^b$ of the background ensemble is
\begin{align}
\vv{P}^b_{j} = (E-1)^{-1}\sum_{i =1}^E (\vv{x}^{b,k}_j - \bar{\vv{x}}^b_j)(\vv{x}^{b,k}_j - \bar{\vv{x}}^b_j)^T,
\end{align}
where $\bar{\vv{x}}^b_j$ is the mean of the background ensemble members. The goal of the analysis step, in which the types of EnKF differ, is to compute an analysis ensemble whose mean $\bar{\vv{x}}^a_j$ and sample covariance $\vv{P}^a_j$ are consistent with the Kalman filter [\onlinecite{Kalman1960new}].

We now present the computations that form the analysis step of the ETKF, as formulated in Ref. \onlinecite{hunt2007efficient} (in which the consistency with the Kalman filter is explained). Recall that $\vv{y}_j$ is a vector of measurements made at time $j\Delta t$.

\begin{enumerate}
\item Create the matrix 
\begin{align}
\vv{X}^b_j &= \left[\quad \vv{x}^{b,1}_j - \bar{\vv{x}}^b_j \quad\vert\quad \vv{x}^{b,2}_j - \bar{\vv{x}}^b_j \quad\vert\quad \dots \quad\vert\quad \vv{x}^{b,E}_j - \bar{\vv{x}}^b_j \quad \right],
\end{align}
whose columns represent deviations of the ensemble members from the ensemble mean.
\item Compute the matrix $\vv{Y}^b_j =  \vv{H}\vv{X}^b_j$ and the vector $\bar{\vv{y}}_j^b$.
\item Create a matrix 
\begin{align}\label{eq:ETKF}
\tilde{\vv{P}}^a_j = \left[(E-1)I/\rho + \vv{C}\vv{Y}^b_j \right]
\end{align}
where $ \vv{C} = (\vv{Y}^b_j)^T\vv{R}^{-1} $ and $\rho$ is the covariance inflation parameter described in Sec.~\ref{sec:dataAssimilation}.
\item Compute $\vv{W}^a_j  = \left[ (E-1) \tilde{\vv{P}}^a_j \right]^{1/2}$, where the $[\cdot]^{1/2}$ denotes the symmetric square root.
\item Compute $\bar{\vv{w}}^a_j = \tilde{\vv{P}}^a\vv{C}(\vv{y}_j - \bar{\vv{y}}^b_j)$ and add it to the $k^{\text{th}}$ column of $\vv{W}^a_j$ to obtain $\vv{w}^{a,k}_j$.
\item Compute $\vv{x}^{a,k}_j = \vv{X}^b_j \vv{w}^{a,k}_j + \bar{\vv{x}}^b_j$. The analysis ensemble is given by $\lbrace \vv{x}^{a,k}_j \rbrace_{1\leq k\leq E}$.
\end{enumerate}

For our results, the analysis ensemble is initialized arbitrarily in a neighborhood of the model attractor. Then, the forecast and analysis steps described above are applied iteratively to generate the analysis time series used to train the ML system.

\bibliography{mletkf}

%
%

%



\end{document}